# Creating a biologically more accurate spider robot to study active vibration sensing

Siyuan Sun[1], Eugene H. Lin[1], Nathan Brown[2], Hsin-Yi Hung[3], Andrew Gordus[3], Jochen Mueller[2], Chen Li[1*]

*Abstract*—Orb-weaving spiders detect prey on a web using vibration sensors at leg joints. They often dynamically crouch their legs during prey sensing, likely an active sensing strategy. However, how leg crouching enhances sensing is poorly understood, because measuring system vibrations in behaving animals is difficult. We use robophysical modeling to study this problem. Our previous spider robot had only four legs, simplified leg morphology, and a shallow crouching range of motion. Here, we developed a new spider robot, with eight legs, each with four joints that better approximated spider leg morphology. Leg exoskeletons were 3-D printed and joint stiffness was tuned using integrated silicone molding with variable materials and geometry. Tendon-driven actuation allowed a motor in the body to crouch all eight legs deeply as spiders do, while accelerometers at leg joints record leg vibrations. Experiments showed that our new spider robot reproduced key vibration features observed in the previous robot while improving biological accuracy. Our new robot provides a biologically more accurate robophysical model for studying how leg behaviors modulate vibration sensing on a web.

## 1. Introduction

Orb-weaving spiders are functionally blind, yet they operate effectively in complete darkness by relying on vibrational cues from their webs [1-9]. Using distinct vibrational signatures, spiders can detect, differentiate, and/or localize prey, predators, mate, offsprings, and abiotic objects [2, 4, 8, 10]. In addition to possessing highly sensitive vibration sensors [11], orb-weaving spiders have evolved diverse behaviors that may facilitate vibration sensing on their webs. Some of these behaviors occur on long timescales (e.g., altering web geometry or tension during web building which lasts typically tens of minutes to ~1–2 h [12-15]), while others happen on short timescales (e.g., dynamic crouching in ~0.5–3 s bursts at ~5–10 Hz [16-17], use legs to pluck the web as brief <1 s or short 1–3 s events [8, 10, 16]). Notably, the rapid recovery phase of crouching is fast enough to excite passive vibrations of the spider–web system afterward, consistent with observations in *U. diversus* and prior robophysical modeling [18]. These behaviors modulate how vibrations travel through the spider–web–target system or even generate vibrations [7] and potentially contribute to vibration sensing. However, despite previous biological and modeling studies on how silk properties, tension, and web architecture modulate frequency content, amplitude, and wave modes of vibration information [6, 7, 11], these works largely focus on passive or simplified conditions. As a result, the physical principles of active vibration sensing—when the spider itself dynamically crouches or plucks—remain poorly understood, in part because it is experimentally difficult to measure vibrations across the entire spider–web–prey system with actively behaving animals [2, 7, 11, 12]. Specifically, experimental studies rely on laser vibrometers that can only measure a single point vibration at a time [5, 12], precluding comprehensive measurements of the entire system with behaving animals [11], whereas simulation studies often rely on finite element analysis to model the web [21], which are computationally too expensive and slow [22] to further include dynamic animal behaviors [2-3, 22].

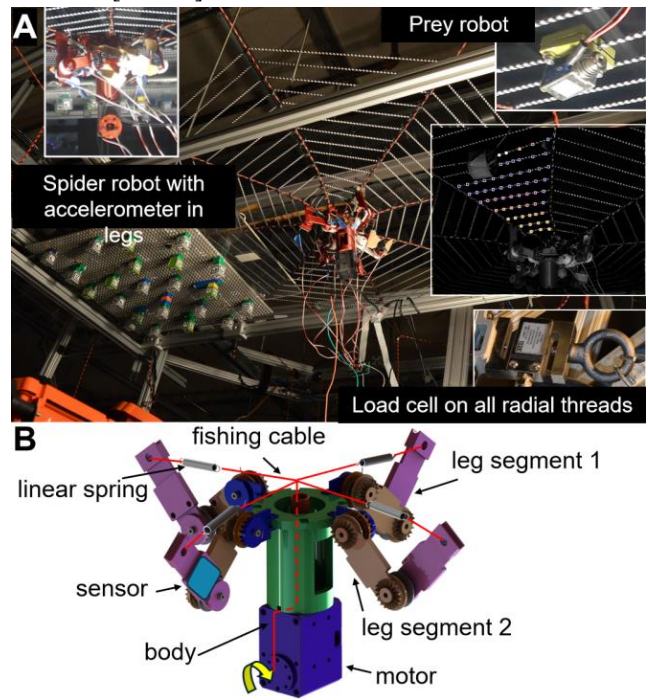

Figure 1. Robophysical model in previous work [18]. (A) Robophysical model of the spider–web–prey system for studying vibration sensing in *U. diversus*. The model features a physical web, a spider robot (198 g) hanging beneath the center of the web, and a prey physical model. Load cells (inset) measure tension in each radial thread of the web. (B) CAD model of the spider robot. A servo motor in the body retracts cables (red) connected to the robot's four legs to actuate their crouching motion (see Fig. 1B). A linear spring embedded in each line allows the leg joints to vibrate. Two

[1]Department of Mechanical Engineering, Johns Hopkins University, Baltimore, MD 21218 USA (e-mail: ▇▇▇▇▇▇▇▇▇▇▇▇▇).
[2]Department of Civil & Systems Engineering, Johns Hopkins University, Baltimore, MD 21218 USA (e-mail: ▇▇▇▇▇▇▇▇▇▇▇▇).
[3]Department of Biology, Johns Hopkins University, Baltimore, MD 21218 USA (e-mail: ▇▇▇▇▇▇▇▇▇▇).

accelerometers on each leg near the joints measure leg vibrations. The distal accelerometer of one leg is shown (cyan).

To address this challenge, our team uses robophysical modeling [23, 24], which enables systematic experiments, enact physical laws, and can generate testable biological hypotheses [13, 23, 24]. Previously, some of us used robophysical modeling to study active sensing of web vibrations to detect prey [22]. This robophysical modeling was based on the orb-weaving spider *Uloborus diversus* [13] as the model organism. This spider constructs a horizontally oriented web, hangs beneath it to capture prey, and dynamically crouches its legs during prey sensing [22]. The robophysical model in the previous work (Fig. 1A) consisted of a spider robot that can crouch its legs, a prey physical robot, and a horizontally oriented physical web made of parachute and shock cords [25].

However, the previous spider robot (Fig. 1B) was highly simplified and not biologically accurate, primarily due to manufacturing constraints. First, the robot only had four identical legs that were radially and symmetrically arranged (Fig. 2C), whereas spiders have eight legs of different lengths with lateral symmetry (Fig. 2A). The choice of having four legs only was made because a single motor could not actuate eight legs simultaneously with the simplified leg design. In addition, each robot leg only had two joints, as opposed to the spider leg's 7 joints. Furthermore, each joint was equipped with a torsion spring such that the entire leg could only flex and extend within a single plane, whereas the spider's legs have more than one degree of freedom at multiple joints and are capable of 3-D leg motion [24]. Finally, the simplistic, cable-driven actuation mechanism (Fig. 1B) only allowed a small range of motion in leg crouching (Fig. 2D) as compared to that of the spider (Fig. 2B).

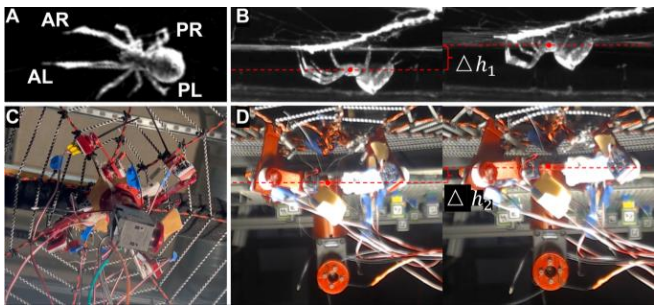

Figure 2. Comparison of the model organism, the orb weaver *U. diversus* and the simplified spider robot in the previous work CITE. (A, C) Bottom view of *U. diversus* (A) and simplified robot (C) hanging under the web. (B, D) Side views of *U. diversus* (C) and simplified robot (D) dynamically crouching its legs from less (left) to more (right) crouched postures. Horizontal dashed lines mark body height at the least and most crouched frames; the resulting Δh (spider: $\Delta h_1$; robot: $\Delta h_2$) shows that, even without normalization, the robot's crouch range is visibly smaller than the spider's.

Despite these limitations, the highly simplified robot proved useful for understanding active vibration sensing via leg crouching. It helped reveal that spiders' leg crouching likely helps detect the presence and distance of prey by inducing prey to passively vibrate at a different frequency, which is anti-correlated with the length of the spiral thread on which the prey is. However, it did not uncover how the direction of prey can be determined, likely because of the inability to differentiate salient spatial patterns of vibrations as it only had four leg each with only two joints. Here, we take the next step in robophysical modeling, by creating a biologically more accurate spider robot (Fig. 3), addressing prior design limitations.

First, to overcome the limitation of having only four identical, radially arranged legs, the new robot incorporates eight legs in a bilateral arrangement, consistent with the morphology of *U. diversus*.

Second, to address the limitation of having only two joints per leg, the new robot design includes four joints per leg, capturing a greater portion of the spider's natural 7-joint limb structure (Sec. 2.1, 2.3). By adding these joints, the robot's legs better approximated spider morphology on the leg level and allowed for a more realistic distribution of motion across the leg segments, providing higher biological fidelity [26].

Third, to enable leg motion in three dimensions, beyond the 2-D sagittal-plane bending of the previous robot, we replaced torsional springs with soft-material silicone joints, in which their geometry were customized to provide some compliance out of the crouching plane while limiting excessive lateral oscillations, allowing combined within-plane crouching and slight out-of-plane lateral motion comparable to real spider legs (Sec. 2.3) [27].

Lastly, to overcome the limited crouching range of motion of the previous robot, we developed a tendon-driven actuation system with pulley-based cable routing (Sec. 2.4). This approach amplified torque and distributed motion across multiple joints, resulting in a greater leg crouching "depth" (range of motion) comparable to that of real spiders. The improvement enables biologically realistic dynamic crouching experiments.

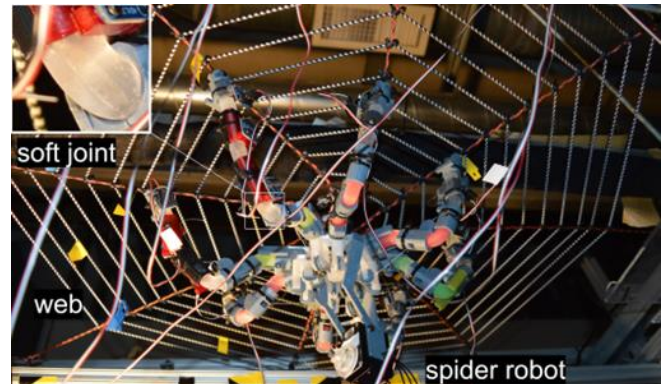

Figure 3. Biologically more accurate spider robot, hanging under the physical web.

## 2. ROBOT DEVELOPMENT

### 2.1. Overall leg design

The leg design of the new spider robot balanced biological accuracy with engineering feasibility by simplifying the spider's seven-joint leg into four-joint structure (Fig. 4A), while prioritizing motion in the sagittal plane that dominates leg crouching behavior during vibration sensing.

Although spider legs possess multiple joints with great multi-axis rotation [28], some adjacent joints, specifically those associated with patella and trochanter (red dots in Fig. 4A), are physically too close to be separated effectively at the scale of our robot. We therefore combined patella segment (pt' in Fig. 4A) and its two adjacent joints into a single effective joint (J2 in Fig. 4C) and merged everything in black box (Fig. 1A) into a single joint (J1 in Fig. 1C).

Because crouching on a web mainly involves sagittal-plane motion, we neglected joint motions that contribute mainly to movements outside the sagittal plane [28], and designed each joint to be flexible mainly along the flexion–extension (pitch) axis. However, we used silicone joints shaped by cross-section geometry and material blends that allow each joint to maintain some flexibility along other directions (Sec. 2.3). This allowed the overall leg to have sufficient flexibility in 3D space so that, when all eight legs were actuated by a single motor (Sec. 2.4), they were not as over-constrained to exert excessive load on the motor as the previous robot.

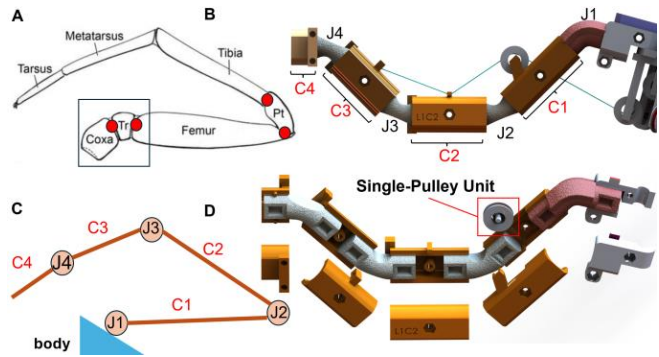

Figure 4. Anatomically informed design of spider robot leg. (A) Schematic of a spider leg, with its natural segments—coxa, trochanter, femur, tibia, metatarsus, and tarsus adapted from [26]. (B) Rendered 3-D model of robot leg. (C) Simplified leg schematic, with 4 leg segments: femur, tibia, metatarsus, tarsus and 4 joints linking adjacent parts. (D) Exploded view showing silicone joints in full and cable routing.

*2.2. Leg geometric proportions*

We further designed each leg to have segmental length proportions that approximate those of the animal's, after considering the simplification mentioned in the previous section. To obtain segment lengths of *U. diversus*' anterior leg (the first pair of legs on each side, Fig. 2A (AL & AR) and posterior leg (the last pair of legs, Fig. 2A (PL & PA), we analyzed side-view high-speed videos of *U. diversus* hanging on its web, recorded at 100 frame/s.

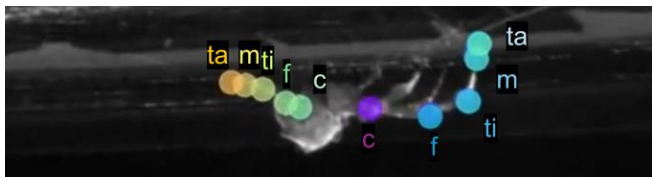

Figure 5. Example side-view of *U. diversus* on its web with manually tracked markers. Colored points mark the five segments of anterior leg (left) and the posterior leg (right): coxa (c), femur (f), tibia (ti), metatarsus (m), and tarsus (ta).

We used DeepLabCut (DLC) [29] to track 5 natural landmarks (coxa, femur, tibia, metatarsus, tarsus) across the anterior left/right and posterior left/right legs (Fig. 2A). Because our analysis only required segment length ratios rather than absolute units, we did not perform a pixel-to-length conversion; all reported values in Table 1 are dimensionless. To obtain these ratios, we normalized femur, tibia, and metatarsus lengths by each leg's tarsus length, assuming that tarsi are short and show minimal variation across legs.

To reduce noise in the video-based measurements, we discarded unreliable points (likelihood < 0.95), linearly interpolated short gaps, and smoothed the trajectories with a zero-phase 4th-order Butterworth low-pass filter (cutoff 60 Hz). For each leg segment, we then calculated the median length across frames. Because side-view DLC only provides 2-D projections, measured lengths are lower bounds on true 3-D segment lengths. During crouching, however, motion is mostly sagittal, so out-of-plane error is modest.

For the two pairs of middle legs, which were often occluded in side-view videos, segment lengths were measured from high-resolution photographs of *U. diversus* obtained from published taxonomic image banks [30]. Measurements were taken in a graphics-annotation tool (Draw), following similar procedures as described above [30]. By normalizing all segment lengths to the tarsus, we assumed that tarsi are effectively equal across legs. This assumption, supported by qualitative visual inspection of published images of *U. diversus*, simplified our analysis, with a small sacrifice on accuracy to capture absolute leg length differences.

Table 1. *U. diversus* leg-segment length ratios. Values are dimensionless segment/tarsus ratios within the same leg (tarsus = 1.0). Medians across frames from side-view DLC (see Methods). For simplification, we used an uniform tarsus length across all legs.

| Leg Pair Index | Femur (C1) | Tibia (C2) | Metatarsus (C3) | Tarsus (C4) |
|---|---|---|---|---|
| Front | 3.1 | 2.83 | 2.5 | 1 |
| Second | 1.7 | 2.36 | 1.55 | 1 |
| Third | 1.4 | 2 | 1.32 | 1 |
| Rear | 2.4 | 2.83 | 1.81 | 1 |

*2.3. Leg joint design*

To allow legs bent in the crouching plane while resisting unwanted lateral motion, we designed silicone joints clamped between 3D-printed connectors (Fig. 6). The connectors had matching grooves and ribs that pressed into the silicone when screwed together, firmly locking the joint in place without glue. This modular assembly allowed quick replacement of joints by loosening the screws, simplifying the testing of different shapes, stiffnesses, or bend angles.

To inform the design of spider leg joints, we reviewed biomechanical and anatomical studies on spider leg morphology [28]. Most spider leg joints possess two or three coupled rotational degrees of freedom—in some cases resembling ball-and-socket or saddle joints—rather than a single hinge axis [31]. Because the torsional springs used in our previous model restricted motion to one plane, we adopted silicone elastomers with tunable properties to better capture the multi-degree-of-freedom flexibility of real spider joints.

The joints were cast in custom 3D-printed PLA molds filled with mixtures of Dragon Skin 0010 platinum-cure silicone rubber (Smooth-On, Inc., Macungie, PA, USA) and Sylgard 182 silicone elastomer (Dow Corning, Midland, MI, USA), then cured in a Heratherm OGS60 gravity convection oven (Thermo Scientific, Waltham, MA, USA) at 50ºC for 3 hrs, removed from the mold, and cured at 100ºC for 1 hr.

Joint cross-sectional geometry was adjusted to limit parasitic motions by increasing the area moment of inertia in the secondary bending direction. During manual bending tests, we observed that the thickened joint design required noticeably different forces when bent along versus across the primary bending axis, suggesting reduced lateral shaking and direction-dependent resistance.

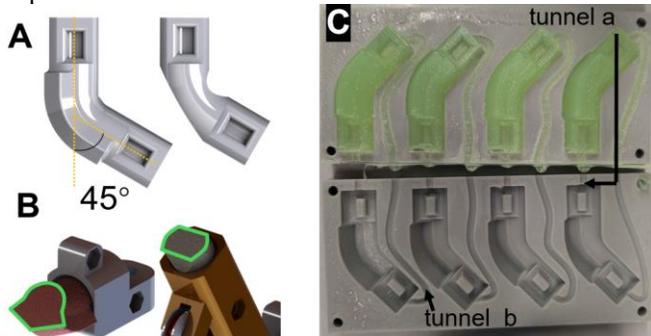

Figure 6. Rendered 3D models for joint design. (A) Side-view comparison of one joint shown with two cross-sectional shapes: thickened type (left) versus regular type (right). Both joints are molded with a 45° bend. (B) Cross-sectional view of the same joints, illustrating the two cross-sectional shapes (thickened vs. regular). (C) Manufacturing sample of four thickened-type 45º joints.

We then varied the joint's cross-sectional shape (standard vs. thickened) to improve stability, material composition (three DS 0010–Syl 182 ratios) to tune stiffness (Table 2), and pre-bent angle (5°, 15°, 30°, 45°, 60°, 75°) to approximate natural spider postures and expand crouching configurations. These three factors—shape, material, and angle—define a design space of 36 joint variants. In this study, we fabricated and tested only a subset, but the modular design provides a versatile platform for future experiments on spider-joint behavior.

Table 2. Material composition ratios (by weight) for robot joints.

| Case | % weight of DS-0010 | % weight of Syl-182 |
|---|---|---|
| 1 | 0 | 100 |
| 2 | 50 | 50 |
| 3 | 70 | 30 |

*2.4. Body design and leg actuation*

To reproduce crouching dynamics observed in previous spider-robot experiments, we designed a centralized actuation system in which a single motor housed in the body simultaneously drives all eight legs to generate coordinated motion.

Specifically, a Dynamixel XM430-W350-R motor (Fig. 7A, C) is housed within the robot body and serves as a central actuation hub for all eight legs. Each leg is connected to the body via a dual-pulley wheel system (Fig. 7B, D) that pairs with a single pulley wheel mounted on the leg itself (Fig. 4D). When the motor rotates a central disk, it simultaneously pulls all eight cables—one per leg—through their respective pulley systems, producing a synchronized crouching motion in the suspended robot (Fig. 7D). Releasing the cable tension allows the legs to passively return to their initial, un-crouched configuration.

To emulate large amplitude dynamic crouching observed in *U. diversus*, we tested and tuned the motor actuation to achieve a large range of motion and high speed for crouching. Although the range of motion increased with angular displacement of the motor, the achievable crouching depth was geometrically constrained by the finite leg length and the initial radial placement of the feet on the web. Specifically, when the feet were placed farther from the web center, a greater fraction of the leg length was required to span horizontal distance, leaving less available length for vertical displacement during crouching. This geometric constraint increased cable tension during deep crouches and risked motor stall (stall torque 4.1 N·m). Based on biological observation, we placed the first pair of feet lay on the 7th spiral thread from the center of the web and the remaining six feet on the 5th spiral thread (Fig. 9A, legs 1 and 5). With this layout, a 360° motor rotation corresponds to the "full depth" crouch while greater rotation produced forces that would stall the motor. This resulted in a slow crouch-recovery cycle (1.2 s), longer than the rapid crouches observed in live *U. diversus* (0.1–0.2 s). To increase crouching speed while maintaining sufficient range of motion, we reduced motor rotation to 180° for all experiments.

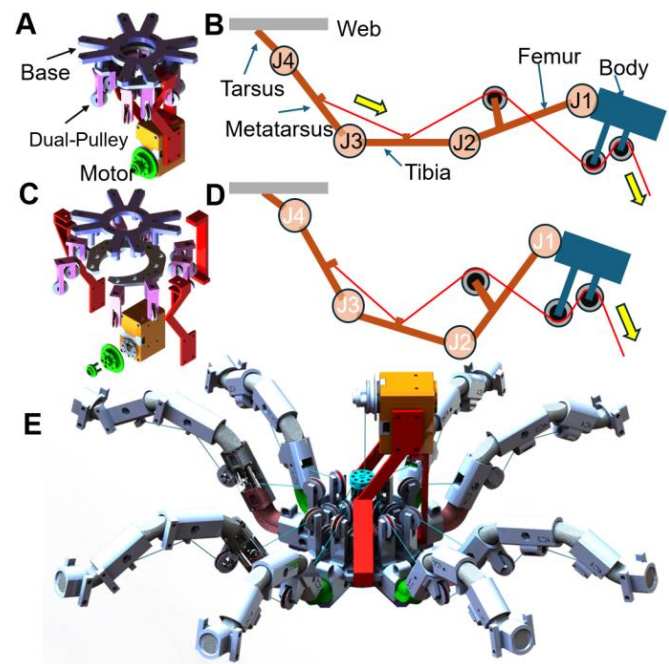

Figure 7. 3-D render of the body design and different views of the complete assembly. (A) Isometric view of the assembly of spider robot's body. (B) Cable slack – relaxed posture. With zero cable tension (red line), all joints remain at their resting angles and the leg's end is fixed to the web; yellow arrows denote directions of applied pulling force generated by the motor. (C) Exploded view highlighting the eight dual-pulley systems. (D) Cable tensioned – crouched posture. Retraction of the cable shortens the red segment, bending all four joints and pulling the leg toward the body into a crouched configuration. (E) Top-oblique views of render of the fully assembled, biologically more accurate spider robot.

We visually compared our robot against *U. diversus*, shown suspended on its web in a posture intermediate between less-crouched and more-crouched states (Fig. 8A, B), reflecting its natural resting position during web-based activities. Here, $\Delta h_{1,2}$ are the same as we defined earlier (Fig. 2B, D), except this time we saw a substantially larger crouching amplitude in the robot (Fig. 8C, D), demonstrating that the new robot achieved a deeper crouching range of motion and closer visual resemblance to the model organism.

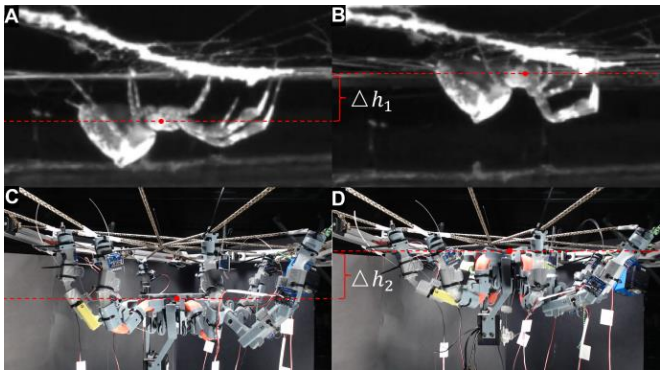

Figure 8. Visual comparison between the model organism *U. diversus* and our spider robot (supplemental video 1). (A, B) Side view of *U. diversus* suspended on its web, displaying a posture intermediate between less crouched and more crouched states. (C) Spider robot hanging on a web in a least crouched state and (D) maximally crouched state (as the robot base nearly touches the web).

## 3. Robot Experiments & Comparison with Previous Work

### 3.1. Experimental protocol

To validate our new robot as a robophysical model for studying vibration sensing during dynamic crouching, we tested whether the robot could differentiate between an empty web and a web containing a prey physical model, following procedures established by our previous work with a simpler spider robot [18]. Specifically, prey presence was represented by a prey physical model (a small mass) attached at the distant end of the web (Figure. 9B)

This passive prey physical model consisted of a small 3D-printed PLA box (60 g, Fig. 10), with compartments for steel plates to vary mass up to 230 g (Fig. 10). The prey mass was intentionally chosen to achieve a biologically relevant prey-to-spider mass ratio. In biological systems, *U. diversus* has a body mass of $(0.0016 \pm 0.0002$ g, mean $\pm$ SEM; n = 24), while typical prey such as *D. melanogaster* have a body mass of $(0.0160 \pm 0.0015$ g, mean $\pm$ SEM; n = 48), corresponding to an approximately one-order-of-magnitude prey-to-spider mass ratio [32]. To approximate this relative loading condition in the robophysical system, we used a prey physical model with a mass up to 230 g relative to the 800 g spider robot. The spider robot was suspended at the center of the web (Fig. 9A) and commanded to perform a single crouch recovery cycle (see Fig. 8C). Leg vibrations were recorded using accelerometers (Adafruit Industries, New York, NY, USA, ADXL326) mounted near metatarsus–tarsus joint of each leg and sampled at 500 Hz via a data acquisition system (USB-231 Measurement Computing Corporation). The crouching event was identified as the period containing a pronounced downward transient in the acceleration signal followed by a return toward the baseline, corresponding to the robot's downward crouch and subsequent recovery. We then low-pass filtered the acceleration signals using a 4th-order Butterworth filter with a cutoff frequency at 25 Hz and applied a fast Fourier transform (FFT) to the crouching segment (0.3–11 s) to obtain frequency spectrum of leg acceleration for each trial. For each condition (web with prey mass and empty web), we performed seven independent crouching trials.

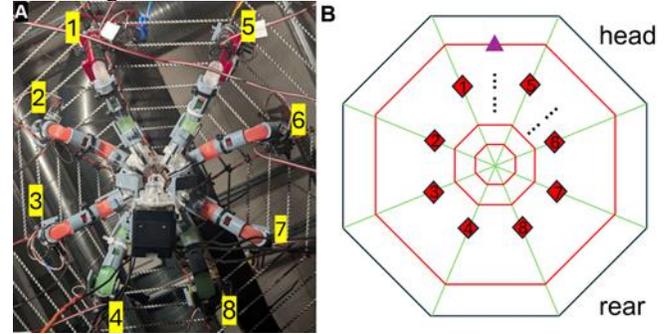

Figure 9. Bottom view of experiment setup. (A) Bottom view of when the robot was secured at the hub (with zip ties), with anterior tarsal tips (leg 1 & 5) fixed at approximately 7th thread and other legs at 5 threads from the hub. (B) Schematic diagram of the spider robot on the web, where red lines are spiral threads and green lines are radial threads. Numbers boxed inside red diamond denote leg number. Purple triangle denotes prey physical model placement which is located on the mid of 11th spiral thread.

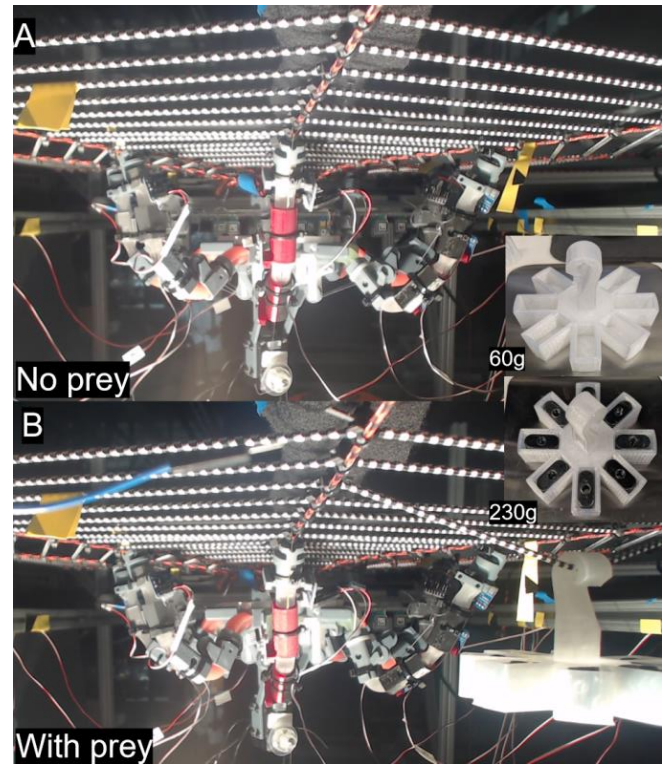

Figure 10. Experiments comparing with and without prey physical model (supplemental video 2 & 3). (A) Spider robot hang on web without prey physical model. (B) Spider robot hang on with prey physical model. Inset shows prey physical model (empty, 60 g vs. fully loaded, 230 g).

We analyzed how dynamic crouching changed the frequency content of vibrations sensed by the robot's legs, building directly on the robophysical analysis in our previous spider-robot study, which showed that dynamic crouching can reveal both the spider robot's natural frequency and an additional prey-dependent frequency in the leg vibration spectra. For each trial, we analyzed the time window containing the crouching motion and the subsequent decay of vibrations, as defined in the experimental protocol (Sec. 3.1). We then computed a single-sided amplitude spectrum

(positive frequencies only) of leg acceleration from the filtered time series over this temporal window. For each leg and treatment condition, we averaged spectra across seven trials and reported the mean ± 1 standard deviation at each frequency to quantify both the typical vibration signature and its trial-to-trial variability.

*3.2. Experimental results*

In the absence of prey physical model, a single dynamic crouch produced damped oscillatory vibrations that were highly similar across all eight legs (Fig. 11A) in which leg accelerations peaked at about +5 to –4 m/s² immediately after the crouch and then decayed over ~2–3 s, leaving only small residual oscillations. The corresponding frequency spectra showed a single dominant vibration peak at ~3.8 Hz for every leg (Fig. 11B), with comparable peak magnitudes around 0.25 m/s² and very little energy outside this band. Trial-to-trial variability in the spectra was small for most legs, indicating that the leg responses were repeatable and largely governed by a common body-web vibration mode in the absence of prey.

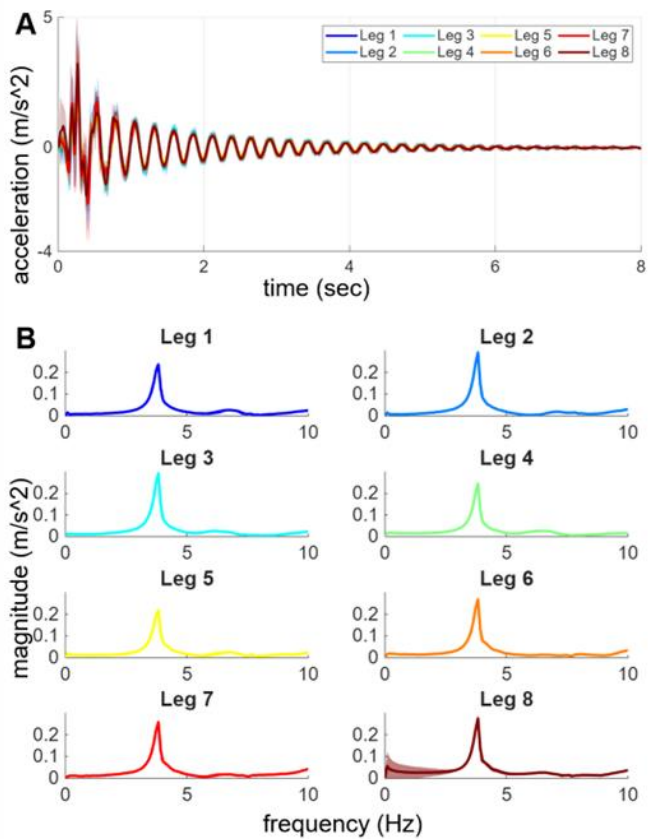

Figure 11. Experimental results without prey. (A) Acceleration vs. time of vibrations across the eight legs. (B) Fourier transform of vibrations across eight legs.

With the prey physical model attached at the web center, a single dynamic crouch again produced damp oscillatory vibrations that were broadly similar across legs in the time domain (Fig. 12A). Peak leg accelerations remained on the order of a few m/s² immediately after crouching, and the vibrations decayed over ~2–3 s, comparable to the no-prey condition.

In the frequency domain, however, the presence of the prey physical model introduced a clear additional feature in the leg vibration spectra (Fig. 12B). All eight legs exhibited a dominant peak at ≈ 3.8 Hz, corresponding to the baseline body–web mode observed in the no prey physical model trials. When the prey physical model was present, a second peak between 5–6 Hz emerged in most legs, especially the middle and posterior legs. Together, these results indicate that the robot can detect prey physical model's presence as an additional, prey-dependent vibration frequency superimposed on the baseline body–web mode.

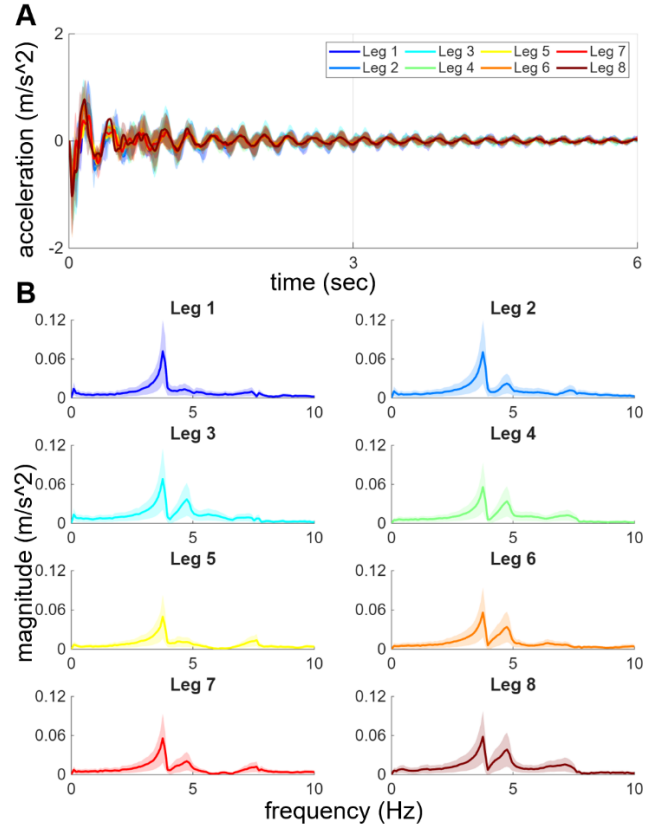

Figure 12. Experimental results with prey physical model. (A) Time-domain plot of filtered acceleration signals for the spider robot's eight legs, measured after a single crouch on the web with a prey physical model attached. (B) Single-sided amplitude spectra for each leg (mean, solid line; ±1 s.d., shaded band). A baseline peak at ≈ 3.8 Hz is present in all legs, while an additional second peak between 5–6 Hz emerges only when the prey physical model is present (most prominently in the middle and posterior legs).

## 4. Discussion

*4.1. Summary of results*

In this study we created a biologically more accurate robophysical spider model that (i) used eight bilaterally arranged legs instead of four radial legs, (ii) increased the number of joints per leg to better approximate spider limb kinematics, (iii) replaced torsional springs with soft silicone joints that allow combined bending and limited lateral motion, and (iv) employed a tendon–pulley actuation system that produces crawling-like multi-joint crouching with a range comparable to real spider. Our experiments demonstrated that this robot can reproducibly generate dynamic crouching on a physical web and detect vibration frequencies across all eight legs of a lower frequency, revealing both a baseline low-frequency mode and an additional higher-frequency component when a prey physical model is present. These vibration features are characteristics of those extensively studied and demonstrated to inform prey physical model's presence and distance as demonstrated in the previous study[22]. These results demonstrated that our new spider robot achieved its design goal of being able to reproduce key features of the earlier robot and improve biological accuracy. This new robot will provide a useful tool for further studying how the spider robot can sense the direction of prey physical model or robot by comparing the sensed leg vibrations across legs.

*4.2 Limitation & future work*

As with any robophysical model, the present spider robot necessarily represents a simplified approximation of the biological system rather than a fully faithful replica. Certain anatomical and mechanical features are idealized or omitted to enable controlled experimentation. For example, the tarsus is modeled as a rigid terminal segment without adhesive pads or fine-scale compliance, despite its known role in contact mechanics and vibration transmission in real spiders. These simplifications allow systematic investigation of core sensing mechanisms but may limit the fidelity of specific aspects of leg–web interaction. Future work should improve the spider robot to further refine the biological accuracy and expand the scope of robophysical studies in several respects.

First, the robot's leg-to-body mass ratio is greater than that of real spiders. In running insects such as the cockroach *Blaberus discoidalis*, all six legs together account for only ~5–6% of total body mass, with most mass concentrated in the body [33]. Similarly, studies of uloborid spiders suggest that most mass is invested in prosoma and opisthosoma, with only a modest fraction in the legs, except for slightly heavier first legs in some species [34]. In contrast, a substantial fraction of the current robot's mass is in the legs (> 50% based on CAD model estimates). This discrepancy could be reduced through targeted redesign and manufacturing choices, including replacing solid leg components with hollow or lattice structures fabricated via multi-material 3D printing, using lower-density exoskeleton materials for distal leg segments, and relocating actuators and ballast closer to the body center to concentrate mass in the prosoma-like region. Together, these changes would reduce leg inertia while preserving joint compliance and kinematics, bringing the mass distribution closer to that of real spiders [22, 32].

Second, although our robot's legs represent a substantial advance over the previous design, they do not yet capture the full mechanical complexity of spider leg joints. Mechanical measurements in spiders such as *Cupiennius salei* and several theraphosid species show that the tibia–metatarsus joint behaves approximately linearly with little hysteresis under axial loading but exhibits strong direction- and rate-dependent viscoelasticity under lateral and dorsoventral loading [35, 36]. In addition, studies of arachnid leg joints indicate that muscle activation, elastic cuticular structures, and internal pressure together shape joint torque, stiffness, and energy storage, implying that effective joint stiffness can be modulated dynamically rather than being purely passive. Similar active tuning of joint stiffness has been demonstrated in insect legs, where co-contraction of antagonistic muscles and passive elastic elements allows stiffness to be regulated independently of joint angle to stabilize limb dynamics and shape mechanical signal transmission[34]. At the scale of our current robot, directly replicating these biological mechanisms remains challenging. However, future work could approximate key aspects of this behavior using soft-robotic design strategies, including multi-material or layered joint constructions to introduce anisotropic stiffness, viscoelastic polymers to capture rate-dependent damping, and miniature compliant actuators or antagonistic tendon arrangements to enable limited active tuning of joint stiffness. Such approaches, which are increasingly explored in miniature soft and compliant robots [36], would enable systematic investigation of how direction-dependent and actively tunable joint mechanics influence vibration transmission and sensing on webs.

Finally, our spider robot currently operates in open loop and therefore cannot yet capture closed-loop crouching behaviors observed in real spiders during prey sensing. Recent work[32] shows that spiders transition between static, crouching, and high-frequency shaking states in a stimulus-dependent manner, with the timing and intensity of crouching modulated by changes in prey-generated vibratory power on the web [33]. While our robot is not designed for locomotion or navigation on the web, future work could incorporate real-time accelerometer feedback to enable closed-loop control of dynamic crouching. Specifically, sensory feedback could be used to modulate the timing, depth, and repetition of crouches in response to measured vibration amplitude or frequency content, enabling systematic tests of how closed-loop crouching strategies shape vibration sensing and prey localization.

## 5. Acknowledgement

We thank Yishun Zhou for constructing the physical web and Yaqing Wang for suggestion on the leg actuation design. This work was supported by an NSF Physics of Living Systems grant (PHY-2310707), co-sponsored by NSF Dynamics, Control and Systems Diagnostics to C.L. and A.G., a Burroughs Wellcome Fund Career Award at the Scientific Interface to C.L.

Author contributions: S.S designed, manufactured, and assembled the robot, wired up data collection tools, conducted experiments, analyzed data, and provided input on the paper. E.L guided the experiments, provided experimental setup and protocol, and data analysis protocol, and revised the paper. N.B and J.M. provided help with manufacturing. H.H and A.G. provided animal data to guide the design. C.L. oversaw the study and revised the paper.